\documentclass[11pt]{article}

\usepackage[final]{acl}

\usepackage{times}
\usepackage{latexsym}
\usepackage{tipa}

\usepackage[T1]{fontenc}
\usepackage[utf8]{inputenc}

\usepackage{microtype}

\usepackage{inconsolata}

\usepackage{graphicx}
\usepackage{ragged2e}
\usepackage{adjustbox}
\usepackage{algorithm}
\usepackage{algpseudocode}
\usepackage{amsmath}
\newcommand\blfootnote[1]{%
  \begingroup
  \renewcommand\thefootnote{}\footnote{#1}%
  \addtocounter{footnote}{-1}%
  \endgroup
}

\title{"Chi nas dal soch el sent de legn" - Auditing Text Corpora for Lombard}

\author{Edoardo Signoroni \\ NLP Centre, Faculty of Informatics \\ Masaryk University \\ Botanická 68a, 602 00 Brno, Czechia \\ \texttt{e.signoroni@mail.muni.cz} \And
Pavel Rychlý\\ NLP Centre, Faculty of Informatics \\ Masaryk University \\ Botanická 68a, 602 00 Brno, Czechia \\ \texttt{pary@fi.muni.cz}}

\begin{document}
\maketitle              % typeset the header of the contribution

\begin{abstract}
Several of the world's languages are still under-resourced in terms of Natural Language Processing (NLP) tools. This is mostly due to the lack of high-quality datasets to train, develop, and evaluate systems and models for several tasks, such as Machine Translation (MT). We conduct a manual audit of the parallel and monolingual corpora available for Lombard, an under-resourced language continuum from Italy. Our analysis reveals that the perceived abundance of web-scraped data is an illusion, with massive datasets plagued by severe language misidentification, boilerplate text, and non-linguistic noise. Furthermore, we analyze the orthographic composition of the valid Lombard portions across web-scraped datasets, curated corpora, and benchmarks. Our findings show conflicting orthographical systems and severe representational bias across all corpora: high-quality data is heavily skewed towards Western Lombard varieties, with Eastern ones left on the margins. This underscores the need for variety-aware, community-driven data curation rather than purely quantity-driven scraping.
\end{abstract}

\section*{Introduction} \label{sec:intro}

\blfootnote{\textit{Chi nas dal soc el sent de legn.} $\rightarrow$ eng: Who's born from the log smells of wood. A saying that could be seen as the Eastern Lombard take on the common phrase "garbage in, garbage out".}

% DATA AS THE BASE
Modern Natural Language Processing (NLP) technologies are built on huge amounts of data, mostly text. One such technology is Machine Translation (MT), which requires parallel corpora of millions of aligned sentences, usually acquired trough automated scraping of the web. These kind of corpora are increasingly available, multilingual, and documented \cite{kreutzer-etal-2022-quality}.

% LOW-RESOURCE
However, of the world's 7000+ languages and variants, only a handful are covered by these efforts. Data scarcity is one of the major issues that so-called under-resourced languages still face in today's NLP. In recent years, several efforts are being made to include these "left-behinds" \cite{joshi-etal-2020-state} in multilingual datasets, such as FLORES+ \cite{nllb-24} or NLLB \cite{nllbteam2022languageleftbehindscaling} and while positive, these inclusion still are not without problems.

% ISSUES WITH DATASETS
As Kreutzer et al. \cite{kreutzer-etal-2022-quality} have shown, these massive, highly multilingual datasets are plagued with severe quality issues such as big portion of the data being non-linguistic noise or in the wrong language. For most of the language pairs in parallel datasets such as CCAligned \cite{el-kishky-etal-2020-ccaligned} and WikiMatrix \cite{schwenk-etal-2021-wikimatrix} the correct examples are less than 50\%, with some direction having none all together. These issues are even more prominent for under-resourced languages.

% ISSUES WITH THIS QUANTITY DRIVEN APPROACH
As the focus for defining under-resourced languages remain on huge datasets \cite{ramponi-2024-language}, some scholars point out the problems of this "quantity over quality" approach. Kreutzer et al. \cite{kreutzer-etal-2022-quality} points to the risks of "representation washing": while these datasets can claim to include under-resourced languages, the lower quality data harms the performance of downstream applications \cite{khayrallah-koehn-2018-impact}, flattens the representation of these languages, and biases the research about them \cite{kreutzer-etal-2022-quality}.

% REPRESENTATIVENESS AND VARIETY AWARE NLP
The representation bias is especially true for corpora of under-resourced languages, which are often assumed to be homogeneous, noise-free images of monolithic language entities and speaker communities. With most of the data for under-resourced languages coming from Wikipedia, there are real risks of creating collections of "wikivarieties" \cite{ramponi-2024-language} made of noise, boilerplate, highly artificial text, and translations from high-resource articles devoid of actual meaningful and culturally relevant content.
Thus, far from being monolithical ISO 639-3 labels, under-resourced and endangered languages are often non-standardized and live as a diverse continuum of related variants for which text may not even be the preferred method of communication \cite{ramponi-2024-language}.

% LOMBARD
All of the above hold true also for the language varieties of Italy, 30 of which are endangered according to the UNESCO \cite{UnescoAtlas}. With recent approaches still struggling, the curation of high-quality, community-driven, and variety-aware datasets is paramount to empower these under-resourced language varieties \cite{signoroni-rychly-2026-llms}. Among these, is Lombard, an endangered Gallo-Romance language variety spoken in the territories around Lombardy, Italy.

% CONTRIBUTIONS and OVERVIEW
This paper sheds light on the current state of the textual corpora for Lombard. Through a manual quality audit of datasets samples and a human-validated orthography variants classification, we show that:
\begin{itemize}
    \item The quality of web-scraped corpora for Lombard is very low, with most of the samples containing less than 25\% of usable examples; curated and benchmark corpora are very good or perfect in terms of alignment, however...
    \item The actual Lombard content of all corpora is a mix of several orthographical systems, but not representative of the diversity of the Lombard continuum, being heavily skewed towards Western Lombard variants.
\end{itemize}

We release our audited samples at \url{repo_removed_for_anonymity.}

\section{Lombard} \label{sec:lombard}

Lombard is a Western Romance language continuum belonging to the Gallo-Italic group, spoken by approximately 3.5 million people primarily in Northern Italy and Southern Switzerland. Despite some internal phonetic, lexical, and morphosyntactic variations, the varieties are largely mutually intelligible \cite{ColuzziPlanning, BonfadiniDialetti, LoporcaroProfilo, ColuzziCase}. Scholars typically categorize the continuum into main branches, most notably Western Lombard and Eastern Lombard. 

Recognized by UNESCO as "definitely endangered" \cite{UnescoAtlas}, and recent statistics highlight a shrinking speaker base, with only 4.2\% of the regional population using it exclusively at home \cite{istat_2026}, even if figures for occasional users may be higher. In fact, Lombard exists in a state of \textit{dilalìa} \cite{berruto_dilalia} with Standard Italian, with the first relegated to mostly informal contexts. 

While Swiss varieties enjoy some institutional support, Lombard in Italy, as is the case with most language varieties, is frequently stigmatized and mislabeled as a mere "dialect" (of Standard Italian) due to complex historical and socio-political reasons. Consequently, modern scholarship prefers the neutral term "language varieties" to avoid prestige-based bias and acknowledge their status as independent linguistic developments \cite{ramponi-2024-language}.

Modern Lombard remains predominantly an oral medium. Despite a rich local literary tradition, the continuum lacks a codified, unified written form. Most speakers employ ad-hoc phonetic spelling or localized orthographies, often tailored for artistic, informal, or highly localized digital communication.

\section{Related Work} \label{sec:related_work}

\subsection{Corpora} \label{sec:corpora_related}

\paragraph{Web-scraped Corpora}
Most of the parallel and monolingual web-scraped corpora we evaluate are accessible through the OPUS collection \cite{tiedemann-opus}. For parallel data, these include WikiMedia \cite{tiedemann-2012-parallel}, WikiMatrix \cite{schwenk-etal-2021-wikimatrix}, and XLEnt \cite{el-kishky-etal-2021-xlent}, a dataset explicitly focused on named entity translations. We also evaluate the Lombard-English portion of NLLB \cite{nllbteam2022languageleftbehindscaling}, one of the largest multilingual parallel datasets currently available. For monolingual data, we analyze recent large-scale web-crawled collections, specifically CulturaX \cite{nguyen-etal-2024-culturax}, Glot500 \cite{imanigooghari-etal-2023-glot500}, and the latest release of the HPLT 3.0 dataset \cite{oepen2025hplt30largescalemultilingual}.

\paragraph{Curated Corpora}
Human-curated datasets are typically smaller, sourced from specific community translation efforts or academic projects, and often hosted on platforms like HuggingFace or GitHub. In our audit, we include the OLDI\_Seed multilingual parallel training dataset \cite{seed-23}, community-driven Minecraft translations \cite{minecraft}, the Piötòst corpus \cite{signoroni2022piotost}, which serves as a manually cleaned and validated subset of WikiMatrix, and GATITOS \cite{jones-etal-2023-gatitos}, a high-quality, multi-way parallel dictionary dataset.

\paragraph{Benchmark Corpora}
Finally, we evaluate standard benchmark corpora used to assess MT performance. We focus our analysis on the current iteration of the FLORES+ evaluation benchmark \cite{nllb-24}, specifically the \texttt{dev} and \texttt{devtest} splits. We omit xP3x \cite{muennighoff2022crosslingual} from our primary analysis, as its Lombard portion essentially mirrors the FLORES+ directions paired with an English instructional prompt.
To our knowledge, no other benchmark datasets exist for Lombard.

\subsection{Annotation} \label{sec:annotation_related}

%QUALITY AT A GLANCE
Kreutzer et al. \cite{kreutzer-etal-2022-quality} conduct a manual audit of the quality of 205 language corpora released as part of five major public datasets. For each corpus and language, they annotate 100 random samples with a simple error taxonomy that can be used also by non-native annotators. Their analysis shows that lower-resource corpora in particular suffer from systematic issues such lack of usable text, low-quality sentences, mislabeled languages, and more. They evaluate the English-Lombard portion of WikiMatrix \cite{schwenk-etal-2021-wikimatrix} to find that only 12\% of the examples are usable correct translations.

%RAMPONI lmo evaluation
Ramponi \cite{ramponi-2024-language} in their wider study on the language varieties of Italy, conducts an audit of samples from CCAligned \cite{el-kishky-etal-2020-ccaligned}, OSCAR \cite{OrtizSuarezSagotRomary2019}, and WikiMatrix \cite{schwenk-etal-2021-wikimatrix}. For Lombard, they find that just 12.8\% of the data is correct, confirming the findings of Kreutzer et al. \cite{kreutzer-etal-2022-quality}.

%piotost evaluation
Signoroni \cite{signoroni2022piotost} also conducts and audit of WikiMatrix \cite{schwenk-etal-2021-wikimatrix}, finding that half of the alignments were completely incorrect.

\begin{table}[htbp]
\centering
\adjustbox{width=\columnwidth}{\begin{tabular}{l|l}
     \textbf{Tag} & \textbf{Meaning} \\
     \hline
     CC & Correct \\
    &  \textit{I polinomi g’hen in tanti aree de la matemategh e di scienzi.} \\
    &  Polynomials appear in many areas of mathematics and science. \\
     \hline
     CS & Correct short \\
     & \textit{Balla de fen} \\
     & Hay bale \\
     \hline
     CB & Correct boilerplate or not useful  \\
     & \textit{(FR) Vist di stazion, insecula.com.} \\
     & (FR) Vedute di stazioni, su insecula.com. \\
     \hline
     X & Wrong Translation \\
     & \textit{perodegh} \\
     & Dorothy Garrod \\
     \hline
     WL & Wrong Language \\
     & \textit{Ma qiáo lì zhuxing wèi "ma hú."} \\
     & The other for yelling, "ho ho ho." \\
     \hline
     NL & Non-Language \\
     & Ïå÷ü 50õ50 ñì, èç 10 - ìì ëèñòà \\
     & Tues, 1/10 - This Is Us \\
     \hline
\end{tabular}}
\caption{The taxonomy proposed by Kreutzer et al. \cite{kreutzer-etal-2022-quality} and used for our audit. Below each tag's explanation, an example from the annotated alignments.}
\label{tab:tagset}
\end{table}

\section{Methodology} \label{sec:methodology}

\subsection{Annotation}

To evaluate the current state of textual resources, we adopt the methodology of Kreutzer et al. \cite{kreutzer-etal-2022-quality}.

For parallel corpora, we focus on Lombard-English and Lombard-Italian parallel corpora. We chose to limit our efforts to these directions, since they are the most probable real-world uses cases for Lombard MT.

The human annotation and evaluation are carried out by a native speaker of \textit{Bresà} (Brescian), an Eastern Lombard variety. For each corpus and direction, we select and manually annotate 100 random examples using the taxonomy detailed in Table \ref{tab:tagset}. 

For the named entity dataset XLEnt \cite{el-kishky-etal-2021-xlent}, we adapted the rules slightly: CB denotes untranslatable copied entities, and CS denotes correctly translated entities. For monolingual corpora, the X (Wrong Translation) tag is omitted. Additionally, when bot-generated templates appeared repeatedly across samples, the first instance was tagged as CC, while subsequent identical instances were penalized and marked as CB (boilerplate).

\begin{table*}[htbp]
    \centering
 \adjustbox{width=\textwidth}{\begin{tabular}{c|c|c|c}
         \hline
         \textbf{Tag} & \textbf{Name} & \textbf{Variant} & \textbf{Grouping} \\
         \hline
         MILCLASS & Classical Milanese & Milanese & Western Classical (WESTCLAS)\\
         NOL & New Lombard Orthography & Pan-Lombard & Western Classical (WESTCLAS) \\
         LOCC & Unified Insubric Orthography & Western Lombard & Western Modern (WESTMOD) \\
         LSI & Ticinese Orthography & Ticinese Lombard & Western Modern (WESTMOD) \\
         LORUNIF & Oriental Unified Orthography & Eastern Lombard & Eastern Modern (EASTMOD) \\
         BERGDUC & Orthography of the Duchy & Bergamasque & Eastern Modern (EASTMOD) \\
         BREMOD & Modern Brescian Orthography & Brescian & Eastern Modern (EASTMOD) \\
         CRES & Cremasque Ortography & Cremasque & Eastern Modern (EASTMOD) \\
         SL & \textit{Scriver Lombard} & Pan-Lombard & Scriver Lombard (SL) \\
         \hline
    \end{tabular}}
    \caption{Summary of Lombard Orthographies and their grouping in our evaluation.}
    \label{tab:ortho_groups}
\end{table*}

% \begin{table}[htbp]
%     \centering
%     \adjustbox{width=\textwidth}{\begin{tabular}{l|l}
%     \hline
%        WESTCLAS  &  \textit{Siddharta Gautama a l'era fi\textbf{oeu} d'ona fame\textbf{j}a ricca e influenta.} \\
%        &    Siddartha Gautama was the son of a rich and influential family. \\
%        WESTMOD  &   \textit{"El v\textbf{öö}r dir che i archivi sun p\textbf{ü}tost diferent di bibliotech per i funziun e l’urganizaziun} [...] \\
%        &    It means that archives are quite different than libraires due to fuctions and organization. \\
%        EASTMOD  &  L'è 'n zèner che g'ha 'mportànsa perchè 'l t\textbf{ö}l dét divèrse spéci cultiàde 'n agric\textbf{ü}lt\textbf{ü}ra [...] \\
%        & It is a genus that has importance because it keeps together several species cultivated in agriculture [...] \\
%        SL  &  \textit{Vergun al di\textbf{x} \textbf{qe} ol Colombo l'hiva visitad la Groenlandia} [...] \\
%        &    Someone says that Columbus has visited Greenland [...] \\
%        \hline
       
%     \end{tabular}}
%     \caption{An example and its translation for each orthography grouping. In \textbf{bold} some of the distinctive markings of each group.}
%     \label{tab:ortho_examples}
% \end{table}

\subsection{Corpora Composition}

We analyze the composition of the Lombard side applying the LombardoGraphia classifiers by Signoroni and Rychlý \cite{signoroni2026lombardographiaautomaticclassificationlombard} on the valid Lombard examples, i.e. the ones not tagged with WL or NL. 
While annotating the quality of the corpora was manageable, in principle, by non-experts, as Kreutzer et al. \cite{kreutzer-etal-2022-quality} found, assigning each Lombard example to a specific orthography was not always possible, even with native knowledge. Lombard lacks a official, recognized written form shared by all variants, thus many proposed competing orthography arose, chosen by personal preference, geographical provenance, and spoken variety. Many examples lacked sufficient graphical or lexical elements to be definitely assigned to one orthography or the other. 

% ORTHOGRAPHY VS VARIANT
Moreover, as pointed out by Signoroni and Rychlý \cite{signoroni2026lombardographiaautomaticclassificationlombard}, in the case of Lombard varieties, orthography and variant cannot be completely disentangled, since they are deeply intertwined and almost overlap in practical use. 

Thus, in our manual annotation, we proceeded as follows: for each example, we applied a rule-based heuristic that excluded non-applicable systems based on specific exclusionary graphemes and lexical choices, e.g. \textit{ç} for \textit{Scriver Lombard} or variations of the word \textit{pussee} ("more"), a distinctively Western Lombard lexeme. We then manually validated a sub-sample of the automatic predictions against the heuristic-driven expert annotation.

This cross-validation confirmed the severe variant under-specification of many of the examples, thus we opted to aggregate the orthography tags in more clearly distinguishable groups: Western Classical (MILCLASS+NOL\footnote{While NOL is conceived as an orthography for all Lombard varieties, its spelling conventions are heavily influenced by MILCLASS and it is mostly used by Western Lombard speakers.}), Western Modern (LOCC+LSI), Eastern Modern (LORUNIF, BERGDUC, BREMOD, CRES), and \textit{Scriver Lombard}\footnote{Similarly to NOL, SL is a Pan-Lombard orthography. Nonetheless, from we could observe from our evaluation, most of its users appear to be Western Lombard speakers.} (SL). Table \ref{tab:ortho_groups} summarizes the orthography variants and their grouping.

\section{Results} \label{sec:results}

\begin{table*}[htbp]
    \centering
\adjustbox{width=\textwidth}{
\begin{tabular}{l|l|c|c|c|c|c|c|c|c|c}
\hline
\textbf{Lang.} & \textbf{Corpus} & \textbf{Ref.} & \textbf{Size} & \textbf{CC+CS} & \textbf{CB} & \textbf{X+WL+NL} & \textbf{WESTCLAS} & \textbf{WESTMOD} & \textbf{EASTMOD} & \textbf{SL} \\
\hline
& \textit{Scraped corpora} \\
\hline
eng & NLLB v1 & \cite{nllbteam2022languageleftbehindscaling} & 8,725,803 & 1\% & 0\% & 99\% & 61\% & 30\% & 9\% & 0\% \\
eng & WikiMatrix v1 & \cite{schwenk-etal-2021-wikimatrix} & 10,435 & 25\% & 6\% & 69\% & 57\% & 32\% & 5\% & 6\% \\
eng & wikimedia v20230407 & \cite{tiedemann-2012-parallel} & 121 & 10\% & 3\% & 87\% & 83\% & 17\% & 0\% & 0\% \\
eng & XLEnt v.1.2 & \cite{el-kishky-etal-2021-xlent} & 19,594 & 2\% & 19\% & 79\% & 64\% & 19\% & 8\% & 9\% \\
ita &  WikiMatrix v1  & \cite{schwenk-etal-2021-wikimatrix} & 11,596 & 25\% & 7\% & 68\% & 62\% & 25\% & 11\% & 3\% \\
ita & wikimedia v20230407 & \cite{tiedemann-2012-parallel} & 346 & 88\% & 2\% & 10\% & 86\% & 4\% & 3\% & 7\% \\
ita & XLEnt v.1.2 & \cite{el-kishky-etal-2021-xlent} & 11,482 & 2\% & 18\% & 80\% & 54\% & 30\% & 6\% & 10\% \\
mono & CulturaX & \cite{nguyen-etal-2024-culturax} & 4,793 & 7\% & 93\% & 0\% & 0\% & 2\% & 98\% & 0\% \\
mono & Glot500 & \cite{imanigooghari-etal-2023-glot500} & 342,156 & 45\% & 23\% & 32\% & 51\% & 32\% & 13\% & 4\% \\
mono & HPLT 3.0 & \cite{oepen2025hplt30largescalemultilingual}& 1,608,399 & 12\% & 1\% & 87\% & 43\% & 48\% & 3\% & 6\% \\
\hline
& \textit{Curated corpora} \\
\hline
eng & GATITOS & \cite{jones-etal-2023-gatitos} & 4,000 & 93\% & 0\% & 7\% & 39\% & 28\% & 7\% & 26\% \\
eng & minecraft-translations & \cite{minecraft} & 7,145 & 73\% & 0\% & 27\% & 73\% & 14\% & 0\% & 13\% \\
eng & OLDI Seed & \cite{seed-23} & 6,193 & 99\% & 1\% & 0\% & 43\% & 31\% & 0\% & 26\% \\
ita & minecraft-translations & \cite{minecraft} & 7,145 & 72\% & 0\% & 28\% & 62\% & 11\% & 1\% & 26\% \\
ita & OLDI Seed & \cite{seed-23} & 6,193 & 100\% & 0\% & 0\% & 41\% & 25\% & 0\% & 34\% \\
ita & Piötòst & \cite{signoroni2022piotost} & 5,306 & 83\% & 6\% & 11\% & 62\% & 24\% & 8\% & 6\% \\
\hline
& \textit{Benchmark corpora} \\
\hline
eng & FLORES+ dev & \cite{nllb-24} & 997 & 100\% & 0\% & 0\% & 21\% & 43\% & 11\% & 25\% \\
eng & FLORES+ devtest & \cite{nllb-24} &1,012 & 100\% & 0\% & 0\% & 17\% & 35\% & 12\% & 36\% \\
ita & FLORES+ dev & \cite{nllb-24} &997 & 100\% & 0\% & 0\% & 17\% & 40\% & 17\% & 26\% \\
ita & FLORES+ devtest & \cite{nllb-24} & 1,012 & 100\% & 0\% & 0\% & 19\% & 41\% & 16\% & 24\% \\
\hline
\end{tabular}
}
    \caption{Annotation results for each corpus. We aggregate usable (CC+CS), boilerplate (CB), and wrong examples (X+WL+NL). The other columns report the aggregated orthography labels. We drop from the orthographical classification examples tagged with WL and NL, however examples tagged as X may still have fluent Lombard.}
    \label{tab:results}
\end{table*}

\subsection{Corpora Quality} \label{sec:quality_results}

Quality varies greatly across corpora typology and direction. 

\paragraph{Scraped Corpora} \texttt{NLLB} claims more than 8.5M Lombard-English parallel examples, however 99\% of our sample is composed by unusable noise. \texttt{WikiMatrix} has around 10k examples for both \textit{lmo-eng} and \textit{lmo-ita}, but in our samples, a quarter is usable, with the remainder being boilerplate or noise. \texttt{WikiMedia} is a tiny corpus of just hundreds of sentences for both directions, with \textit{lmo-eng} being very noisy (12\% of the sample tagged as CC+CS). The \textit{lmo-ita} sample is instead quite clean, with almost nine out of ten examples being usable. Our sample of \texttt{XLEnt} shows that it fails almost completely to align entities for both directions: 2\% are actual translated entities, a further 18-19\% are copied names or untranslatable toponyms.

Monolingual corpora are not much better in terms of quality: just 7\% of our \texttt{CulturaX} sample is indeed usable text and, while there are no outright wrong examples, 93\% of the sample is just boilerplate text, almost all bot-generated Wikipedia stubs for towns and public transport stations. \texttt{Glot500} seems to be the most useful monolingual corpus, with 45\% of the sample being acceptable text. There were several cases of examples tagged as WL, mostly being car advertisements in Romanian. Only 12\% of the \texttt{HPLT 3.0} sample is correct, and again most of the noise consisted of language misidentification: instead of Lombard, the sample contains several instances of adjacent language varieties such as Romansh, Ladin, and Piedmontese. As Kreutzer et al. \cite{kreutzer-etal-2022-quality} suggests, advancements in language identification, especially for similar variants and dialects \cite{aepli-etal-2022-findings}, may ameliorate the issue. Otherwise, without a way to automatically separate valid examples from noise, using the data is impractical, arguably rendering the whole corpus unusable.

\paragraph{Curated Corpora}

Curated corpora have, as expected, an overall higher quality.
In our sample of \texttt{GATITOS} almost all alignments were correct, but mostly short due to the nature of the dataset. Moreover, while \texttt{GATITOS} has multiple translation alternatives for one English example, in some instances the translation correct, but quite arbitrary and narrow in scope: e.g. eng: match $\rightarrow$ lmo: \textit{solfanell} ("matchstick") instead of \textit{partìda} ("sports match") or \textit{paregià} ("to match things together"); or eng: can $\rightarrow$ lmo: \textit{toella} ("tin can") instead of \textit{pudì} ("can", "to be able to").\footnote{The examples from \texttt{GATITOS} are in WESTCLAS or SL, while we give ours in EASTMOD.}
For both directions, \texttt{minecraft-translations} is composed of around 70\% correct, but mostly short and very specific, translations. The rest is mostly untranslated WL, primarily Italian. 
Our sample confirms that in terms of alignment quality \texttt{OLDI\_Seed} is, as expected, the gold standard for both \textit{lmo-eng} and \textit{lmo-ita}, with just one CB example in the \textit{lmo-eng} direction and no other noise.
\texttt{Piötòst} is mostly clean, but our sample shows that some 10\% erroneous alignments still persist after the human validation. The correct examples, however, most probably overlap with its noisy "parent" corpus, \texttt{WikiMatrix}, to which this curated corpus is a cleaner version.

\paragraph{Benchmark Corpora} The \texttt{FLORES+} \texttt{dev} and \texttt{devtest} benchmark dataset are perfectly aligned regarding translations.

\subsection{Corpora Composition} \label{sec:composition_results}

While looking at the orthographical composition of the corpora, it is evident that the data is heavily skewed towards Western Lombard conventions, primarily the "classical"-inspired ones. Only \texttt{CulturaX} seems to be almost all written according to Eastern orthographies, however, as we have shown, this corpus is composed almost exclusively of boilerplate text. In practice, actual text in Eastern varieties is almost non-existent in available datasets.

At a first glance, only benchmark datasets appear to be more representative, however, this does not stand to a closer inspection: Western systems still account for around 60\% of the data, while only around 15\% is written in Eastern orthographies. 

While almost absent from scraped corpora, SL takes up more space in both curated and benchmark datasets. Even if it was conceived as system for all Lombard variants, lexical cues in our dataset suggest that is is mostly used by Western Lombard speakers, further aggravating the linguistic biasing of the datasets.

Further complicating the issue, even in the benchmark data there were some instances of examples features from different orthographic systems, e.g. SL \textit{qe} and WESTMOD \textit{ü}, which is not part of the SL set.

% LANGUAGE LABELS AS MONOLITHS vs VARIETY AWARE
These findings confirms what Ramponi \cite{ramponi-2024-language} and Kreutzer et al. \cite{kreutzer-etal-2022-quality}, among others, already suggested: the Lombard in these datasets is presented as uniform and clean, without further validation. Even in the structured environment of Wikipedia or curated projects such as OLDI Seed or FLORES+, the diversity of the Lombard continuum seeps in, leading to conflicting, fictitious, or distorted representations of each variant and of Lombard as a whole.

Table \ref{tab:results} summarizes the results of our annotation.

\section{Conclusions} \label{sec:conclusions}

In this paper, we presented a comprehensive manual audit of the textual corpora available for the Lombard language continuum. Our evaluation confirms and expands previous research to confirm that, especially for under-resourced varieties such as Lombard, web-scraped corpora consist almost entirely of noise, misidentified languages, or repetitive boilerplate text. 

Furthermore, while human-curated datasets and benchmarks offer highly accurate alignments, they suffer from a severe representational bias. They mix several orthographical conventions, especially Western Lombard ones, leaving Eastern varieties virtually absent from usable data. 

To effectively empower the speakers of under-resourced language continuums, the NLP community must move beyond blind, quantity-driven web scraping. Future efforts must prioritize advanced, dialect-aware language identification and invest in targeted, community-driven data curation that respects and captures internal linguistic variation to create data, benchmarks, and tools that work with and for the speakers communities.

\section*{Limitations}

This study relies on the manual evaluation of a single Eastern Lombard native speaker annotator. Due to the wide range of orthographical systems proposed for Lombard, for some of them the annotation confidence may be lower.

Also for this reason, and due to the severe variant under-specification of the text, we had to conflate several distinct orthographies into broader macro-groups. This approach still leaves much to be desired regarding representativeness, and this is why we advocate for a community-driven approach where all speakers can apply their knowledge and variants can be properly represented.

\section*{Ethical Concerns}

Developing NLP tools for endangered, fragmented languages carries the risk of digital homogenization. The overwhelming bias toward Western Lombard in current training and benchmark datasets risks artificially elevating one specific variety to a "standard" digital status. If deployed in downstream generative or translation applications, this bias could inadvertently contribute to the erasure of the linguistic heritage and diversity of speakers of other varieties, counteracting the restorative goals of low-resource NLP.

\section*{Acknowledgements}
The work described herein has been supported by the Ministry of Education, Youth and Sports of the Czech Republic, Projects No. LM2023062 LINDAT/CLARIAH-CZ and PRINS CZ.02.01.01/00/23\_025/0008710.

\bibliography{custom}

\end{document}